\documentclass[a4paper]{article}
\usepackage{Odyssey2018}
\usepackage{epsfig,amssymb,amsmath}
\usepackage{multirow}
\usepackage{booktabs}
\usepackage{url}
\usepackage[pdftex]{color}
\newcommand{\cprm}{$C^{\rm Prm}_{\rm min}$}

\ninept

\title{Gaussian meta-embeddings for efficient scoring of a heavy-tailed PLDA model}

\makeatletter
\def\name#1{\gdef\@name{#1\\}}
\makeatother
\name{{\em Niko Br\"ummer$^1$, Anna Silnova$^2$, Luk\'a\v{s} Burget$^2$ and Themos Stafylakis$^3$}}
\address{
1. Nuance Communications, South Africa \\ 
2. Brno University of Technology, Czech Republic\\ 
3. Computer Vision Lab, University of Nottingham \& Omilia Conversational Intelligence, UK \\
{\small \tt niko.brummer@gmail.com, isilnova@fit.vutbr.cz} }

\def\R{\mathbb{R}}

\def\detm#1{\lvert#1\rvert}
\def\logdet#1{\log\detm{#1}}

\def\dot#1#2{\expv{#1,#2}{}}

\def\Lset{\mathcal{L}}
\def\Rset{\mathcal{R}}

\def\Sset{\mathcal{S}}

\def\zvec{\mathbf{z}}
\def\avec{\mathbf{a}}
\def\rvec{\mathbf{r}}
\def\cvec{\mathbf{c}}

\def\nulvec{\boldsymbol{0}}
\def\onevec{\boldsymbol{1}}
\def\etavec{\boldsymbol{\eta}}

\def\Imat{\mathbf{I}}
\def\Bmat{\mathbf{B}}
\def\Gmat{\mathbf{G}}
\def\Wmat{\mathbf{W}}

\def\Fmat{\mathbf{F}}
\def\Lambdamat{\mathbf{\Lambda}}
\def\Gammamat{\mathbf{\Gamma}}

\def\expvb#1#2{\left\langle#1\right\rangle_{#2}}
\def\expv#1#2{\bigl\langle#1\bigr\rangle_{#2}}

\def\ND{\mathcal{N}}
\def\TD{\mathcal{T}}

\begin{document}
\maketitle

\begin{abstract}
Embeddings in machine learning are low-dimensional representations of complex input patterns, with the property that simple geometric operations like Euclidean distances and dot products can be used for classification and comparison tasks. We introduce meta-embeddings, which live in more general inner product spaces and which are designed to better propagate uncertainty through the embedding bottleneck. Traditional embeddings are trained to maximize between-class and minimize within-class distances. Meta-embeddings are trained to maximize relevant information throughput. As a proof of concept in speaker recognition, we derive an extractor from the familiar generative Gaussian PLDA model (GPLDA). We show that GPLDA likelihood ratio scores are given by Hilbert space inner products between Gaussian likelihood functions, which we term Gaussian meta-embeddings (GMEs). Meta-embedding extractors can be generatively or discriminatively trained. GMEs extracted by GPLDA have fixed precisions and do not propagate uncertainty. We show that a generalization to heavy-tailed PLDA gives GMEs with variable precisions, which do propagate uncertainty. Experiments on NIST SRE 2010 and 2016 show that the proposed method applied to i-vectors without length normalization is up to 20\% more accurate than GPLDA applied to length-normalized i-vectors.

\end{abstract}

\section{Introduction}
\emph{Embeddings} are familiar in modern machine learning. Neural nets to extract word embeddings\footnote{\url{en.wikipedia.org/wiki/Word_embedding}} were already proposed in 2000 by Bengio~\cite{Bengio_word_embedding}. Now embeddings are used more generally, for example in state-of-the-art face recognition, e.g.\ Facenet~\cite{Facenet}.

Embeddings are becoming popular also in speech and speaker recognition. At Interspeech 2017, eighteen papers had the word `embedding' in the title.\footnote{\url{www.interspeech2017.org/program/technical-program/}.} In speaker recognition and spoken language recognition, we have been using \emph{i-vectors}---embeddings extracted by a generative model---for almost a decade~\cite{ivector-Brighton,ivec,BUT_ivector_language}. More general embeddings extracted by discriminatively trained DNNs are now appearing in speaker recognition, see for example the \emph{OK Google} system~\cite{end2end_google}, the \emph{Voxceleb} paper~\cite{Voxceleb} and JHU's \emph{x-vectors}~\cite{end2end,DSIS17,DS_ICASSP18,SRI_xvector}. Similar embeddings are also being used for spoken language recognition~\cite{LIMSI_Language_embedding,Jesus_Odyssey2018,JHU_Odyssey2018}. 
 
Input patterns (sequences of acoustic feature vectors, images, text, \ldots) live in large, complex spaces, where probability distributions and geometric concepts such as distance are difficult to formulate. The idea with embeddings is that they are representations of complex input patterns that live in simpler spaces, e.g.\ $\R^d$ (multidimensional Euclidean space), where distance is naturally defined and can be put to work to compare patterns. 

At the Johns Hopkins HLTCOE SCALE 2017 Workshop\footnote{\url{http://hltcoe.jhu.edu/research/scale/scale-2017}} the ongoing research on embeddings for speaker recognition~\cite{DS_ICASSP18} inspired the generalization to \emph{meta-embeddings}. The bulk of the work on meta-embeddings remains unpublished, but a current draft of that work can be followed on GitHub~\cite{meta_embeddings}.

Traditional embeddings can be interpreted as point estimates for hidden variables of interest and they typically live in low-dimensional Euclidean spaces, where comparisons between them are based on ordinary dot products. Meta-embeddings are \emph{likelihood functions} for those hidden variables. They (meta-embeddings) typically live in infinite-dimensional Hilbert function spaces and comparisons between them are based on more generally defined inner products. This is a considerable generalization, which provides many new opportunities, but also complex challenges, both theoretical and computational. In this work, we restrict attention to multivariate Gaussian likelihood functions, for which the required inner products can be evaluated in closed form. 

In future we hope to apply meta-embeddings in speaker recognition in a similar way to i-vectors or x-vectors, in the sense that they will be extracted from the acoustic feature vectors (MFCCs). We regard the work in this paper as a warm-up exercise and a proof on concept, in which we use i-vectors, rather than MFCCs as input. With i-vector inputs, we can profit from simple generative models (Gaussian or heavy-tailed PLDA) that provide elegant closed-form formulas for extracting meta-embeddings.

\section{Motivation}
In speaker recognition, we already have generative (i-vector) and discriminative (x-vector) embeddings that represent the state of the art and beyond. What additional advantages could we expect from meta-embeddings? The main motivation is to re-design from the ground up a vehicle for the propagation of uncertainty. 

The generative i-vector extractor model~\cite{ivec} provides a natural measure of uncertainty in the form of the i-vector posterior. In standard i-vector scoring recipes (PLDA~\cite{HTPLDA} and cosine scoring~\cite{ivector-Brighton}), only the expected value of that posterior is retained, while the precision (inverse covariance) is discarded. Work to propagate this source of uncertainty through PLDA has had limited success~\cite{Uncertainty-Sandro, Uncertainty-Patrick, Uncertainty-Themos}. According to Patrick Kenny~\cite{Uncertainty-Bilbao}, i-vector uncertainty propagation does not do what it is supposed to do, but instead gives some benefit as a channel compensator. We speculate that simplifying modelling assumptions and the mean-field variational Bayes approximation (which is required to make the i-vector extractor tractable) may be factors contributing to this problem.

To our knowledge, publications on discriminatively extracted embeddings for speaker recognition have not yet addressed the issue of uncertainty propagation. \footnote{Although the discriminative x-vector extractor does make use of standard deviations in its temporal pooling stage~\cite{DS_ICASSP18}, the uncertainty thus captured is not propagated through to the subsequent PLDA scoring backend.}

As we show below, meta-embeddings, whether extracted discriminatively or generatively, are designed to propagate uncertainty. Let us motivate this in a more general pattern recognition context. Quantifying the uncertainty is very important if a pattern recognizer is to be applicable to variable and sometimes challenging conditions. In speaker recognition, a short, noisy, narrow-band recording should leave much more uncertainty about the speaker than a long, clean, wideband recording. In face recognition, compare a well-lit, high resolution, full-frontal face image to a grainy, low resolution, partly occluded face. In fingerprint recognition, compare a clean, high-resolution ten-print, to a single, distorted, smudged fingermark retrieved from a crime scene.  
 
\section{Meta-embeddings}
For an elaborate tutorial introduction to meta-embeddings, the interested reader is encouraged to read the first four chapters of~\cite{meta_embeddings}. Our summary here is limited to a few paragraphs.

In very general terms, we can describe speaker recognition as the problem of partitioning sets of recordings according to speaker~\cite{SPP}. Set sizes can vary from entire databases to \emph{binary trials} that contain just a pair of recordings. For simplicity we assume: each recording has a single speaker; recordings from different speakers are independent; and the recordings of a given speaker are exchangeable. By De Finetti's theorem~\cite{Chow}, exchangeability is equivalent to the concept of the \emph{hidden speaker identity variable}, which is familiar in speaker recognition thanks to the work of Patrick Kenny in JFA~\cite{JFA} and PLDA~\cite{HTPLDA}. Meta-embeddings are \emph{likelihood functions} for the speaker identity variable, of the form: 
$$f(\zvec)\propto P(r\mid\zvec)$$
where $\zvec$ is the identity variable and $r$ denotes some representation of a recording, e.g.\ raw speech, MFCCs, i-vector, etc.

For traditional embeddings, the intuitive idea is to retain in the output representation as much as possible of the relevant information that was present in the input, $r$. For meta-embeddings, that idea is formalized: by definition \emph{all} of the relevant information about the speaker of $r$ must be present in the likelihood function of that recording. (Keep in mind that information content has meaning only relative to some probability model~\cite{PTLOS} and that for a poor likelihood model, asserting we have all of the relevant information won't do us any good in practice! Just as in any other probabilistic machine learning task, we need to choose the likelihood models wisely and find ways of training the parameters of these models.)

Let $\Rset=\{r_j\}_{j=1}^n$ denote a set of $n$ recordings. \footnote{In this paper, the $r_j$ are i-vectors, but in future work they will be sequences of MFCCs.} In this paper, we let $\zvec\in\R^d$ denote a $d$-dimensional hidden speaker identity variable, for which we assign the standard normal distribution as prior: $\pi(\zvec)=\ND(\zvec\mid\nulvec,\Imat)$. The \emph{meta-embedding}, $f_j$, extracted from input $r_j$ is the likelihood function:
\begin{align}
\label{eq:me_def}
f_j(\zvec) &= k_j P(r_j\mid\zvec)
\end{align}
where $k_j>0$ is an arbitrary constant that may in general depend on $r_j$. Take careful note: the meta-embedding is the \emph{whole function}, $f_j$, rather than some point estimate that lives in $\R^d$. 

Let $A,B$ represent two different hypotheses of how $\Rset$ might be partitioned w.r.t.\ speaker. Partition $A$ has $m$ hypothesized speakers, indexed by the subsets, $\{\Sset_i\}_{i=1}^m$, where $\Sset_i\subseteq\{1,\ldots,n\}$. Likewise, $B$ has $m'$ speakers, indexed by $\{\Sset'_i\}_{i=1}^{m'}$. Using within-speaker exchangeability and between-speaker independence, the likelihood ratio (LR) comparing $A$ to $B$ can be expressed in terms of the meta-embeddings, $f_j$ as:\footnote{This formula is a variation of the principle of \emph{Q-scoring}, which we have introduced in~\cite{q_scoring,Jesus_Odyssey2018}.}
\begin{align}
\label{eq:generalLR}
\begin{split}
\frac{P(\Rset\mid A)}{P(\Rset\mid B)} &=
\frac{\prod_{i=1}^m \expvb{\prod_{j\in\Sset_i} P(r_j\mid\zvec)}{\pi}}
{\prod_{i=1}^{m'} \expvb{\prod_{j\in\Sset'_i} P(r_j\mid\zvec)}{\pi}} \\
&= \frac{\prod_{i=1}^m \expvb{\prod_{j\in\Sset_i} f_j(\zvec)}{\pi}}
{\prod_{i=1}^{m'} \expvb{\prod_{j\in\Sset'_i} f_j(\zvec)}{\pi}} 
\end{split}
\end{align}
where the triangle brackets denote expectation w.r.t. $\pi(\zvec)$. The arbitrary scaling constants, $\{k_j\}_{j=1}^n$, are the same in the numerator and denominator and cancel. This equation is very general: any speaker recognition problem that can be formulated in terms of partitions can be expressed purely in terms of the meta-embeddings. This illustrates the principle that the likelihoods represent all the relevant information in the inputs. 

If we look only at the first line of~\eqref{eq:generalLR}, we might conclude that to recognize speakers with a principled probabilistic model, we would always need a generative model, which requires possibly complex probability distributions for the observed data, of the form $P(r_j\mid\zvec)$. Keeping in mind that $f_j(\zvec)$ is essentially an un-normalized posterior for $\zvec$:
$$P(\zvec\mid r_j)\propto \pi(\zvec) f_j(\zvec)$$
the last line of~\eqref{eq:generalLR} shows however that we require only much simpler distributions for $\zvec\in\R^d$. 

Although we \emph{can} extract meta-embeddings from generative models (as we do below), the generative models are by no means required. The RHS of~\eqref{eq:generalLR} shows that we can score---and therefore also train---meta-embedding systems in purely discriminative ways, without requiring complex generative models for the input data. 
\begin{quote}
\emph{This paves the way for discriminatively trained, DNN-based speaker recognizers, with principled uncertainty propagation.}
\end{quote}

\subsection{LR, inner product and pooling}
For a binary trial, when $\Rset=\{r_1,r_2\}$, there are only two possible hypotheses, $H_1$: there is one speaker; and $H_2$: there are two speakers. Let $f_1$ and $f_2$ denote the meta-embeddings extracted from $r_1,r_2$, as defined by~\eqref{eq:me_def}. The likelihood ratio~\eqref{eq:generalLR} simplifies to:
\begin{align}
\label{eq:LR}
\begin{split}
\frac{P(\Rset\mid H_1)}{P(\Rset\mid H_1)} &= \frac{\expv{f_1(\zvec)f_2(\zvec)}{\pi}}{\expv{f_1(\zvec)}{\pi}\expv{f_2(\zvec)}{\pi}}\\
&=\frac{\dot{f_1}{f_2}}{\dot{f_1}{\onevec}\dot{f_2}{\onevec}}
\end{split}
\end{align}
where we have defined the constant function, $\onevec(\zvec)=1$ as well as the \emph{inner product} between two meta-embeddings as:
\begin{align}
\dot{f_1}{f_2} &= \int_{\R^d} f_1(\zvec) f_2(\zvec) \pi(\zvec) d\zvec
\end{align}
We further need the concept of \emph{pooling}. Let $\{r_j\}_{j=1}^k$, be a number of recordings all assumed to be from the same speaker and let $\{f_j\}_{j=1}^k$, be the associated meta-embeddings. Then the \emph{pooled meta-embedding} is the product 
\begin{align}
\bar f(\zvec)&=\prod_{i=1}^k f_i(\zvec)
\end{align}
This is the likelihood function conditioned on those $k$ recordings, which is the un-normalized form of the pooled posterior:
\begin{align}
P(\zvec\mid \{r_j\}_{j=1}^k) &\propto \pi(\zvec) \bar f(\zvec)
\end{align}
As shown in~\cite{meta_embeddings}, all likelihood ratios of the form~\eqref{eq:generalLR} can be expressed in terms of these two primitive operations: pooling and inner products. Since our inner products are expectations of products, we can alternatively let our primitive operations be \emph{pooling and expectation}.

Given some regularity conditions on the likelihood functions, our meta-embeddings live in a Hilbert space, which is a vector space equipped with an inner product. Although this Hilbert space is typically infinite-dimensional, it has a geometry just like Euclidean space. In meta-embedding space, norms, distances and angles are well-defined and have meaningful interpretations and practical applications in scoring and training of speaker recognizers. We lack space here, but the details are in~\cite{meta_embeddings}.

\subsection{Gaussian meta-embeddings}
In practice, we need our primitive operations (pooling and expectation) to be computationally tractable. In this paper we restrict attention to multivariate Gaussian likelihood functions. The \emph{Gaussian meta-embedding} (GME), extracted from a recording $r_j$ is defined as:
\begin{align}
f_j(\zvec) &= \exp\bigl[\avec'_j\zvec-\frac12\zvec'\Bmat_j\zvec\bigr]
\end{align} 
where $f_j$ is represented by its \emph{natural parameters}: $\avec_j\in\R^d$ and the $d$-by-$d$ positive semi-definite \emph{precision matrix}, $\Bmat_j$. In future work, we envisage a discriminatively trained meta-embedding extractor where a DNN would process $r_j$ (a sequence of MFCCs) and then output $\avec_j$ and some sensible representation of $\Bmat_j$. In this paper, as warm-up exercise, we let the $r_j$ be i-vectors and we use the PLDA model to derive relatively simple functions to extract $\avec_j,\Bmat_j$.

Since Gaussians are closed w.r.t.\ products and our representation is essentially logarithmic, pooling is easy: we simply need to add the natural parameters. The toolbox of primitive operations is completed by a closed-form expression for Gaussian expectations. For a (raw or pooled) meta-embedding, $f$, represented by $\avec,\Bmat$, the expectation w.r.t.\ the prior is~\cite{meta_embeddings}:
\begin{align}
\label{eq:Edef}
\begin{split}
\log E(\avec,\Bmat) &= \log\expv{f}{\pi} = \log \int_{\R^d} \avec'\zvec-\frac12\zvec'(\Bmat+\Imat)\zvec\,d\zvec \\
&= \frac12\avec'(\Bmat+\Imat)^{-1}\avec - \frac12\logdet{\Bmat+\Imat}
\end{split}
\end{align}
In the general case, Cholesky factorization would be the standard tool for this computation, but the GMEs that we extract in this paper have diagonalizable precisions that allow for much faster computations. 

We can combine the pooling and expectation formulas to compute arbitrary likelihood ratios of the form~\eqref{eq:generalLR}. As an example, for binary trials, the LR becomes:
\begin{align}
\label{GME_binary_LR}
\frac{P(\Rset\mid H_1)}{P(\Rset\mid H_2)} &= \frac{E(\avec_1+\avec_2,\Bmat_1+\Bmat_2)}{E(\avec_1,\Bmat_2)E(\avec_2,\Bmat_2)}
\end{align}

\section{PLDA as meta-embedding extractor}
\label{sec:PLDA}
\def\barB{\bar\Bmat}
\def\Gmat{\mathbf{G}}
In what follows, we let our recording representations be $D$-dimensional i-vectors: $\rvec_j\in\R^D$, and we derive a meta-embedding extractor via a heavy-tailed PLDA model. For more details see the chapter on \emph{Generative meta-embeddings} in~\cite{meta_embeddings}.

The hidden variable is $\zvec\in\R^d$ and we require: $d<D$. The standard normal prior, $\pi(\zvec)=\ND(\zvec\mid\nulvec,\Imat)$, forms part of our PLDA model. (This is a simplification of Kenny's original model, which had a heavy-tailed prior~\cite{HTPLDA}.) The PLDA model says that for every speaker, the identity variable $\zvec$ is sampled independently from $\pi(\zvec)$; and every i-vector, $\rvec_j$, for a speaker having identity variable $\zvec$, is generated as:
\begin{align}
\rvec_j &= \Fmat \zvec + \etavec_j, &
\etavec_j &\sim \TD(\nulvec,\Wmat,\nu)
\end{align}
where $\Fmat$ is a $D$-by-$d$ factor loading matrix and where the `channel' noise, $\etavec_j$, is drawn from a multivariate t-distribution~\cite{PRML}, having zero mean,\footnote{We can let the model have non-zero mean, but in practice it is simpler just to zero the mean of the training data.} precision $\Wmat$ and \emph{degrees of freedom}, $\nu>0$. The PLDA model parameters are $\Fmat,\Wmat,\nu$. The meta-embedding that this model extracts from $\rvec_j$ is simply:
\begin{align}
f_j(\zvec) &\propto \TD(\rvec_j \mid \Fmat\zvec, \Wmat,\nu)
\end{align}
This is a t-distribution for $\rvec_j$, but what does it look like as a function of $\zvec$? Provided that $D>d$ and $\Fmat'\Wmat\Fmat$ is invertible, it is shown in~\cite{meta_embeddings} that it is another t-distribution, with \emph{increased} degrees of freedom, $\nu'=\nu+D-d$
\begin{align}
f_j(\zvec) &\propto \TD(\zvec\mid \Bmat_j^{-1}\avec_j,\Bmat_j,\nu')
\end{align} 
where
\begin{align}
\label{eq:Banda}
\Bmat_j &= b_j \barB, &
\avec_j &= b_j \Fmat'\Wmat\rvec_j, &
b_j &= \frac{\nu+D-d}{\nu+\rvec'_j\Gmat\rvec_j}
\end{align}
and
\begin{align}
\label{eq:barBandG}
\barB &= \Fmat'\Wmat\Fmat, &
\Gmat &= \Wmat-\Wmat\Fmat\barB^{-1}\Fmat'\Wmat
\end{align}
In a typical PLDA model, we have $d\in [100,200]$ and $D\in [400,600]$, so that $\nu'=\nu+D-d$ is large, making the meta-embedding practically Gaussian. We therefore approximate:
\begin{align}
\label{eq:SGME_approx}
f_j(\zvec) &\approx \ND(\zvec\mid \Bmat_j^{-1}\avec_j,\Bmat_j^{-1}) 
\propto \exp\bigl[\avec'_j\zvec-\frac12\zvec'\Bmat_j\zvec\bigr]
\end{align}
Several comments are in order:
\begin{itemize}
	\item For the heavy-tailed case, with small $\nu$, the meta-embedding precisions, $\Bmat_j=b_j\barB$, vary as a function of the data. This is the uncertainty propagation.
	\item Since precisions differ only by a scalar, we can simplify pooling and expectation. For pooling, addition of precisions simplifies to scalar addition. For the expectations~\eqref{eq:Edef}, we do not have to keep Cholesky factorizing every time. By precomputing an eigenanalysis of $\barB$, we can diagonalize all $\Imat+\Bmat_j$, which allows fast scoring and training~\cite{meta_embeddings}.
	\item Notice that $\Gmat\Fmat=\nulvec$, so that 
	$$\rvec'_j\Gmat\rvec_j=\etavec'_j\Gmat\etavec_j$$ 
	is an ancillary statistic, \emph{independent} of the speaker variable $\zvec$, but nevertheless important for complete inferences about the speaker. The heavy-tailed noise, $\etavec_j$ is essentially Gaussian noise with random variance. For a `bad' i-vector, with a large noise vector,  $\rvec'_j\Gmat\rvec_j$ will also be large and the precision scaling factor, $b_j$ will be small. For a `good' i-vector, with small noise, the opposite happens.
	\item For the more familiar case of Gaussian PLDA, when $\nu\to\infty$, we have $b_j\to 1$ and the meta-embedding precisions remain constant. In this case we cannot exploit the informtation in $\rvec_j'\Gmat\rvec_j$.
\end{itemize}

\subsection{Length normalization}
In~\cite{HTPLDA} heavy-tailed PLDA was shown to be a better model of i-vectors than Gaussian PLDA, but the computational cost was considerable. Subsequently,~\cite{Dani_length_norm} showed that the i-vectors could instead be Gaussianized via a simple length normalization procedure. This matched the accuracy of heavy-tailed PLDA, with negligible extra computational cost. The heavy-tailed nature of i-vectors was also addressed in~\cite{Sandro_Bilbao}, where an iterative scaling procedure was proposed.

In this paper, our approximation~\eqref{eq:SGME_approx} has the advantage of a fast, closed-form estimate of the scale factor: $\rvec_j'\Gmat\rvec_j$. Of course, length normalization would interfere with that estimate. We therefore apply the proposed Gaussian meta-embeddings (GMEs) to i-vectors \emph{without} length normalization.  In our experiments below we compare against Gaussian PLDA applied to i-vectors both with and without length normalization.

\subsection{GME extractor and scoring}
In summary, the PLDA model (whether Gaussian or heavy-tailed), provides the functional form~\eqref{eq:Banda}, \eqref{eq:barBandG} and~\eqref{eq:SGME_approx}, for extracting Gaussian meta-embeddings from i-vectors. We shall explore both generative and discriminative methods for training the parameters of this GME extractor. 

It is also worth emphasizing the following difference between i-vectors and x-vectors on the one hand, versus meta-embeddings on the other:
\begin{itemize}
	\item The extractors for i-vectors and x-vectors are typically trained separately from the PLDA scoring backend. (In practice, this has many advantages.)
	\item Meta-embeddings, once extracted, do not need an additional backend for scoring. The meta-embeddings contain within themselves everything that is needed to produce scores and this is done in general by~\eqref{eq:generalLR}, or more specifically for GMEs applied to binary trials by~\eqref{GME_binary_LR}.
\end{itemize}
Of course, i-vectors and x-vectors \emph{can} be scored in a parameterless backend via cosine scoring, but this is usually less accurate.

In machine learning, e.g.\ in Facenet~\cite{Facenet}, the parameterless scoring strategy is preferred. The philosophy there can be summarized as: Train the embedding extractor such that embeddings of the same face are close (in Euclidean distance), while embeddings of different faces are far apart. That brute-force, geometric strategy has been very successful and should be appreciated as such. We do however hope that additional benefits can be eventually reaped as a result of the probabilistic strategy of the meta-embedding design.

Finally, it is worth repeating that comparisons between meta-embeddings \emph{can} be interpreted in terms of a slightly more complex geometry, in terms of angles and norms. The interested reader is referred to the chapter entitled \emph{The structure of
meta-embedding space} in~\cite{meta_embeddings}.

\section{Training}
\def\Zmat{\mathbf{Z}}
In what follows, let $\Rset$ denote the recordings in the training database; $\Lset$, the `labels', or true partition of the database w.r.t.\ speaker; $\Zmat$ all of the hidden speaker identity variables (one per speaker); $\theta$ the generative model parameters (in this paper $\theta=(\Wmat,\Fmat,\nu)$; and $\phi$ a set of variational parameters which would be needed for variational Bayes (VB) training strategies. 

Training strategies for complex probabilistic models are perhaps best understood via comparison with the celebrated \emph{variational autoencoder} (VAE)~\cite{VAE} as described next. 

\subsection{Generative training}
VB training effects approximate maximization of the marginal likelihood---here $P(\Rset\mid\Lset,\theta)$. Classical VB maximizes a lower bound to the marginal likelihood and is applicable to conjugate-exponential models with intractable posteriors and marginal likelihoods~\cite{PRML}. The VAE generalizes this to a wider class of (deep generative) models by using a \emph{stochastic approximation} of the VB lower bound~\cite{VAE}. The VAE has two parts, decoder and encoder. The decoder is the generative model likelihood, $P(\Rset,\mid\Lset,\Zmat,\theta)$. The encoder is a tractable variational posterior, $Q(\Zmat\mid\Rset,\Lset,\phi)$, which approximates the true intractable posterior, $P(\Zmat\mid\Rset,\Lset,\theta)$. VAE training is accomplished by maximization of the stochastic VB lower bound w.r.t.\ both decoder and encoder parameters, $\theta$ and $\phi$. 

In this paper, we did not use VAE. Instead we used a crude shortcut. We trained a Gaussian PLDA model with the usual EM-algorithm and then simply plugged in a hand-selected value for $\nu$ to make the model heavy-tailed. 

Future work may however investigate VAE as a more powerful generative training strategy. The heavy-tailed PLDA model of section~\ref{sec:PLDA} has intractable posteriors, $P(\zvec\mid r_j)$, formed by the product of the Gaussian prior and the t-distribution likelihoods. As we pointed out however, the t-distribution likelihood is almost Gaussian, so that Gaussian variational posteriors can be expected to work well. Unfortunately, the VB lower bound (Gaussian expectations of logarithms of t-distributions) does not have a closed form, so that VAE rather than classical VB would be necessary. For this model, the variational parameters could be tied to the generative parameters, using~\eqref{eq:Banda} and~\eqref{eq:barBandG}. However, VB allows unconstrained optimization of the variational parameters, which could give advantages in both accuracy and computational complexity.

Alternatively, another solution for training is the classical mean-field VB solution for heavy-tailed PLDA of~\cite{HTPLDA}, where the channel noise scaling factors are treated as hidden variables.

\subsection{Discriminative training}
\label{sec:discr_train}
VAE training might be a good idea for simpler models such as heavy-tailed PLDA applied to i-vectors. However, for more complex models applied to acoustic features, discriminative training starts to look more attractive. 
\begin{itemize}
	\item For VAE training, we have a complexity doubling effect---we have to build a complex encoder, a complex decoder and also manage the non-trivial interface between them---see for example~\cite{q_scoring}. Once training is complete, the decoder is no longer needed for runtime scoring. 
	\item In discriminative training, no decoder is needed and we only need to train the equivalent of the encoder.
\end{itemize}
In future work, for more complex models that extract meta-embeddings from acoustic features, we envisage that purely discriminative training methods could provide an easier route to success. 

\emph{Binary cross-entropy} (BXE), applied to pairs of recordings, is a popular discriminative training criterion in speaker recognition~\cite{STBUFusion,BurgetL_ICASSP:2011,calibration_Odyssey14,end2end_google,end2end}. BXE can indeed also be used to train meta-embedding extractors and that is what we do in this paper. The scoring formula needed during BXE training of the GME extractor is~\eqref{GME_binary_LR}.\footnote{Although we did not do that experiment here, notice that discriminative training of the GME extractor with $\nu\to\infty$ is equivalent to discriminatively trained Gaussian PLDA~\cite{BurgetL_ICASSP:2011,Sandro_pairs}.}

For future work, we note that BXE is by no means the only option---see~\cite{meta_embeddings} for a variety of other proposed discriminative training criteria. In particular, we would like to highlight the computationally attractive \emph{pseudolikelihood} criterion, which does not rely on a quadratic expansion into binary trials. In addition, pseudolikelihood is a \emph{proper scoring rule} for this training problem in a stricter sense than BXE and may give advantages as a calibration-sensitive training and evaluation criterion.

\section{Experiments}

\subsection{I-vector extraction}
In this paper our recordings are represented by i-vectors. For all of the experiments we use a single database of i-vectors, extracted as described below. 

We used $60$-dimensional spectral features: $20$ MFCCs, including $C_0$, augmented with $\Delta$ and $\Delta\Delta$. The features were short-term mean and variance normalized over a $3$ second sliding window. With those features, we train a GMM UBM with $2048$ diagonal components in gender independent fashion. Then, we collect sufficient statistics and train the i-vector extractor, with $600$-dimensional i-vectors. These i-vectors serve us as input to either the PLDA baseline or to the new meta-embedding extractor. In both cases, i-vectors are transformed with global mean normalization. Then, for one of the baseline PLDA systems (baseline 1) we also apply length normalization. The second PLDA system (baseline 2) is applied to i-vectors without length normalization.

\subsection{Datasets and evaluation metrics}
UBM, i-vector extractor and PLDA models are trained on the PRISM dataset~\cite{PRISM}, containing Fisher parts 1 and 2, Switchboard 2, 3 and Switchboard cellphone phases. Also, NIST SRE 2004--2008 (from the MIXER collections) are added to the training. In total, the set contains  approximately 100K utterances coming from 16241 speakers. We used 8000 randomly selected files for UBM training and full set to train i-vector extractor and PLDA.

For training the GME extractor we use the same training list. However, for discriminative training via stochastic gradient descent (SGD), we split the set into training and cross-validation subsets. Cross validation was done on a randomly selected subset of 10\% of the speakers, leaving the other 90\% for training. This gave 8740 utterances for cross validation and 90309 for training. 

We evaluate performance on the female part of NIST SRE 2010, condition 5, which consists of English telephone data \cite{NISTSRE10_evalplan}. Additionally, we report the results on the NIST SRE 2016 evaluation set (both males and females). We report the results on the whole SRE'16 and also separately for each of the two language subsets, Cantonese and Tagalog.
As evaluation metrics, we use the equal error rate (EER, in \%) as well as the average minimum detection cost function for two operating points ($C^{\rm Prm}_{\rm min}$). The two operating points are the ones of interest in the NIST SRE 2016 \cite{NIST_SRE_2016}, namely the probability of target trials  $0.01$ and $0.005$.

\subsection{GME extractor initialized from PLDA}
\label{sec:plda_init}
As mentioned in section~\ref{sec:PLDA}, the GME extractor defined by~\eqref{eq:Banda} with $\nu\to\infty$ is equivalent to Gaussian PLDA. In this case, we can set the parameters of the GME extractor in such a way that log-likelihood scores computed with meta-embeddings will be equal to the scores provided by Gaussian PLDA. To see this, recall that in Gaussian PLDA the log-likelihood-ratio score for two i-vectors, $\rvec_1$ and $\rvec_2$, can be expressed as
\begin{equation}
\label{eq:pldacsore}
S_{\mathrm{PLDA}}=2\rvec_1'\Lambdamat \rvec_2+\rvec_1'\Gammamat \rvec_1+\rvec_2'\Gammamat \rvec_2 +(\rvec_1+\rvec_2)'\cvec+k,
\end{equation}
where the parameters $\Lambdamat$, $\Gammamat$, $\cvec$ and $k$ are calculated from the parameters of the PLDA model (see eq. (8) of \cite{BurgetL_ICASSP:2011}). Since we subtracted the global mean from the i-vectors, we have $\cvec=\nulvec$, so that the linear term can be omitted.

Substituting for $\avec$ and $\Bmat$ in \eqref{eq:Edef} with expressions given by \eqref{eq:Banda} and setting $b_j=1$, the GME log-likelihood ratio score is:
\begin{equation}
\label{eq:gmescoring}
\begin{aligned}
S_{\mathrm{GME}}&=\frac{1}{2}\rvec_1'\Wmat'\Fmat((\Imat+2\barB)^{-1}-(\Imat+\barB)^{-1})\Fmat'\Wmat\rvec_1\\
&+\frac{1}{2}\rvec_2'\Wmat'\Fmat((\Imat+2\barB)^{-1}-(\Imat+\barB)^{-1})\Fmat'\Wmat\rvec_2 \\
&+\rvec_1'\Wmat'\Fmat((\Imat+2\barB)^{-1})\Fmat'\Wmat\rvec_2 \\
&-\frac{1}{2}\log|\Imat+2\barB|+\log|\Imat+\barB|
\end{aligned}
\end{equation}
Now, comparing \eqref{eq:pldacsore} and \eqref{eq:gmescoring}, we see that $S_{\mathrm{GME}}=S_{\mathrm{PLDA}}$ when
\begin{equation}
\label{eq:params_to_init}
\begin{aligned}
\Gammamat&=\frac{1}{2}\Wmat'\Fmat((\Imat+2\barB)^{-1}-(\Imat+\barB)^{-1})\Fmat'\Wmat,\\
\Lambdamat&=\frac{1}{2}\Wmat'\Fmat((\Imat+2\barB)^{-1})\Fmat'\Wmat,\\
k&=-\frac{1}{2}\log|\Imat+2\barB|+\log|\Imat+\barB|.
\end{aligned}
\end{equation}
Our GME extractor was initialized by solving for $\Fmat$ and $\Wmat$, while various values for $\nu$ were plugged in by hand. In the heavy-tailed regime (small $\nu$), our results are not very sensitive to the exact value---we report results for $\nu=2$. As a sanity check, we also tried $\nu\to\infty$ to verify the equivalence with Gaussian PLDA. This is achieved by setting the $b_j=1$.

In our experiments below, we try the generatively initialized GME extractor as-is, as well as a discriminatively trained extractor as described next.

%If we are able to set parameters of meta-embeddings ($\Fmat$ and $\Wmat$) such that~\eqref{eq:params_to_init} is true, we can be sure that GME model will perform exactly the same as PLDA from which it was initialized. 
%%%shall i give exact details on how to do that?
%As we said before, to be able to exactly reproduce PLDA model with Gaussian meta-embedding we have to take $\nu\to\infty$. That implies that precision scaling factor $b$ is the same for all i-vectors and equal to 1. Or, in other words, all meta-embeddings have same fixed precision, $\barB$, and there is no propagation of uncertainty. Now, we can try to set $\nu$ to some relatively small value, this way precision scalar stops being independent on i-vector and could be calculated as in \eqref{eq:Banda} for every GME. Our hope is that there exists such value for $\nu$ that would allow current GME (equivalent to Gaussian PLDA) to account for i-vector uncertainty. 

\subsection{Discriminative GME extractor training}
As mentioned in section~\ref{sec:discr_train} we used binary cross-entropy (BXE) scored on pairs of i-vectors to discriminatively train the parameters of the GME extractor. We used the initialization from Gaussian PLDA as described above, with $\nu$ plugged in, followed by minibatch stochastic gradient descent (SGD) on the BXE objective. 

The (rather large) minibatches are formed as follows. We randomly select\footnote{with replacement} two sets of $5000$ i-vectors each from the training data to serve as enrollment and test sets. Then, all i-vectors in the first set are scored against all in the second. We do not filter out any trials except for the cases where i-vectors are scored against themselves. As expected, the amount of non-target trials in a batch is much higher than target trials and the ratio can vary between the batches. To compensate for this, we separately compute BXE for target and non-target examples in the current batch and normalize each individual term by correct number of targets and non-targets respectively. 

The extractor parameters, $\Wmat$ and $\Fmat$, are updated by backpropagating gradients through the BXE objective, through the scoring formula~\eqref{GME_binary_LR} and the extractor formula~\eqref{eq:Banda} and~\eqref{eq:barBandG}. The value of $\nu$ remains fixed at the plugged in value throughout training. Training continues until the BXE objective stops improving on the held out cross-validation set. 

\subsection{Multi-enroll trials}
The SRE'16 evaluation set includes some trials with multiple enrollment recordings. For the results reported here, we took the shortcut of simply averaging enrollment i-vectors. Of course, both PLDA and meta-embeddings provide for more principled enrollment pooling and that will be explored in future.

\subsection{Results}
Table~\ref{tbl:results} compares the results for the two Gaussian PLDA baselines against three variants of Gaussian meta-embeddings.

The first part of the table shows the results of our two baselines. Baseline 1 is Gaussian PLDA with $D=600$ and $d=200$, applied to length-normalized i-vectors. Baseline 2 is the same, but without length normalization. As usual, length normalization helps a lot---it makes the data more Gaussian to better fit the Gaussian PLDA model. 
%\begin{table}
%\caption{\label{tbl:baselineres} Results of baseline PLDA systems trained on length normalized i-vectors (baseline 1) and i-vectors with no normalization (baseline 2) in terms of \cprm and EER, reported in $[\%]$.}
%\vspace{3mm}
%\centerline{
%\begin{tabular}{l c c c c c }
%\toprule
%\multirow{2}{*}{System}   &  \multicolumn{2}{c}{SRE10,c05} & & \multicolumn{2}{c}{SRE16,eval}\\
%\cmidrule{2-3}
%\cmidrule{5-6}
%  & \cprm  & EER & & \cprm  & EER\\
%\midrule
%baseline 1 &  0.262 &  2.54 & &  0.959 & 16.66 \\
%baseline 2 &  0.329 &  4.21 & &  0.963 & 19.13 \\
%\bottomrule
%\end{tabular}}
%\end{table}

The second part of the table shows results for three GME configurations, all of them applied to i-vectors \emph{without} length normalization. The first GME result is the sanity check that shows the equivalence between Baseline 2 and the GME extractor initialized from it, with $\nu\to\infty$. The second GME result is the same as the previous one (PLDA initialization, no further training) but now with $\nu=2$. Notice that this already does better in all cases than Gaussian PLDA without length normalization. Finally the third GME result shows that after further discriminative training, \emph{GME without length normalization can do better than Gaussian PLDA with length normalization}.

By changing the degrees of freedom parameter, $\nu$, we have effectively relaxed the Gaussian modelling assumptions. In our experiments, we have tried several different values for $\nu$. We found that parameter $\nu$ can vary in a wide range of values where all of them provide similar performance. Here, we picked $\nu=2$. As results indicate, the training not only mitigates the degradation brought by the lack of length normalization but even brings further improvements compared to both baselines in most of the cases. 

\begin{table*}
\caption{\label{tbl:results} Comparison of accuracies on SRE 2010 and 2016 of Gaussian PLDA with length normalization (baseline 1) and without it (baseline 2), versus GME (without length normalization).}
\vspace{3mm}
\centerline{
\begin{tabular}{l c c c c c c c c c c c c}
\toprule
\multirow{2}{*}{System}   &  \multicolumn{2}{c}{SRE10 c05,f} & & \multicolumn{2}{c}{SRE16, all} & & \multicolumn{2}{c}{SRE16, Cantonese}& &\multicolumn{2}{c}{SRE16, Tagalog}\\
\cmidrule{2-3}
\cmidrule{5-6}
\cmidrule{8-9}
\cmidrule{11-12}
  & \cprm  & EER & & \cprm  & EER & & \cprm  & EER & & \cprm  & EER\\
  \midrule
baseline 1 &  0.262 &  2.54 & &  0.959 & 16.66 & & \bf{0.684} & 9.80 & & 0.983 & 21.53\\
baseline 2 &  0.329 &  4.21 & &  0.963 & 19.13 & & 0.726 & 13.03 & & 0.985 & 23.10 \\
\midrule
%GME $\nu\to\infty,\;b=1$&   0.329 &  4.21 & &  0.963 & 19.13 & & 0.726 & 13.03 & & 0.985 & 23.10 \\
GME $\nu\to\infty$, no training &   0.329 &  4.21 & &  0.963 & 19.13 & & 0.726 & 13.03 & & 0.985 & 23.10 \\
GME $\nu=2$, no training &  0.299 &  2.87 & &  0.960 & 16.99 & & 0.692 & 10.03 & & 0.979 & 21.63 \\
GME $\nu=2$, retrained & \bf{0.213} &  \bf{2.05} & &  \bf{0.879} & \bf{15.56} & & 0.734 & \bf{9.51} & & \bf{0.955} & \bf{20.66}\\
\bottomrule
\end{tabular}}
\end{table*}

\subsection{Computational complexity}
In our experiments, starting from i-vectors, Gaussian PLDA scoring of the whole SRE'10 and SRE'16 evaluation sets required respectively $0.4s$ and $1.5s$ in wall clock time. The GME solution required roughly double the time for the same tasks. We did not have an implementation of Kenny's heavy-tailed PLDA~\cite{HTPLDA} to hand for direct comparison, but we do know that its computational complexity has thus far prevented it from being widely adopted.

\section{Discussion}
This paper introduces \emph{meta-embeddings}, which are intended as a future alternative to i-vectors and x-vectors in speaker recognition, and indeed in other areas of machine learning as an alternative to traditional embeddings. The chief motivation for meta-embeddings is to build discriminatively trainable recognizers that allow the principled propagation of uncertainty, all the way from the input to the final output. We expect these advantages to be most noticeable in applications with varying and sometimes challenging quality of the inputs.

We do not yet have a full meta-embedding replacement for i-vectors or x-vectors, but were able to demonstrate the utility of our new design principles by creating a new i-vector scoring backend that is more accurate than the long-standing state of the art represented by length normalization and Gaussian PLDA. For SRE'10, we showed a $20\%$ relative improvement in EER and for SRE'16 a $1\%$ absolute improvement. This improvement was achieved purely by replacing the backend---without resorting to data augmentation, fusion, domain adaptation, or score normalization.

Our ongoing work on meta-embeddings can be followed on GitHub at~\cite{meta_embeddings}.

\subsection{The shrinkage problem}
Although our results show that heavy-tailed PLDA performs better than Gaussian PLDA on i-vectors without length normalization, there remains a problem with i-vectors extracted from recordings with short durations. In the usual i-vector extractor, the effect of the standard normal prior is to shrink short-duration i-vectors towards the origin. The heavy-tailed PLDA model breaks down in such cases, because  when $\rvec_j'\Gmat\rvec_j$ decreases, it extracts meta-embeddings with \emph{higher} precision. This is the opposite of what we want---shorter durations should give more uncertainty, not less. This inconsistency can be explained as follows. In an ideal world, the whole PLDA model should form the prior for the i-vector extractor. This is however not practical and we are forced to compromise by using the simpler standard normal prior instead. 

We conjecture that the $\pi$-vectors of~\cite{pi-vec} might help to mitigate this problem. The $\pi$-vectors are extracted similarly to i-vectors, but without the regularizing prior and do not shrink towards zero for short durations. Another way to see it is that $\pi$-vectors are point-estimates extracted from the i-vector likelihood function, rather than from the i-vector posterior. The likelihood function is uncontaminated by the inaccurate prior. 

This is however not a completely satisfactory solution, because we are still going via a point-estimate, which does not properly convey the uncertainty inherent in short-duration recordings. As mentioned, our next goal is to construct meta-embedding extractors that work directly on the acoustic features, rather than via i-vector-like point-estimates.

\section{FAQ for future research}
\def\xvec{\mathbf{x}}
\def\yvec{\mathbf{y}}
\textbf{Q:} I have an existing DNN that extracts x-vectors from MFCCs, followed by a Gaussian PLDA backend. How do I generalize this to a meta-embedding extractor?

\noindent\textbf{A:} One solution is to replace your PLDA with our new GME backend. Replace our $\rvec_j$ with your x-vectors. Initialize the backend from PLDA (with $D \gg d$) as we did. It is probably not worthwhile learning $\nu$, just fix it to say 1 or 2. Discriminative training of the GME backend (with fixed extractor) may already improve accuracy, as it did for us. The next step is to \emph{jointly} optimize the extractor and the backend. This should encourage the uncertainty to be propagated from the extractor to the backend, through the $(D-d)$-dimensional complement of the speaker subspace.

\noindent\textbf{A:} Variants of the above recipe, where the interface between the original extractor and the back-end is simplified may ultimately give better solutions, but they may be more difficult to initialize. For example, one could replace~\eqref{eq:Banda} and~\eqref{eq:barBandG} as follows. Split your $D$-dimensional x-vector into two parts: say $\xvec_j\in\R^d$ and $\yvec_j\in\R^{D-d}$. Then do $b_j=\frac{\nu+D-d}{\nu+\yvec'\yvec}$, $\avec_j=b_j\xvec_j$ and $\Bmat_j=b_j\Lambdamat$, where $\Lambdamat$ is a trainable diagonal matrix with positive entries.\\

\noindent \textbf{Q:} Which discriminative training criterion should I use? 

\noindent\textbf{A:} In our experience, the multiclass cross-entropy that is currently used to train the Kaldi x-vector extractors of~\cite{DS_ICASSP18} and~\cite{SRI_xvector} will not work for training the GME backend, and by implication also not for the joint training stage. Do use your existing method to pre-train your extractor, but then for the further training, change the criterion. The first option to try is BXE as used in this paper. We were not successful in using BXE to train from random initialization, but it did work after PLDA initialization. A look at pseudolikelihood, or some of the other criteria proposed in~\cite{meta_embeddings} may be worthwhile. \\

\noindent \textbf{Q:} Are there any important tricks that are not mentioned in the paper?

\noindent \textbf{A:} Yes. For example, to simplify the backend, set $\Wmat=\Imat$ and learn a linear transform of the $\rvec_j$ instead.  Constrain (or coerce with a suitable L2 regularization penalty) $\barB = \Fmat'\Fmat$ to be diagonal. We would be happy to assist with further details.

\noindent \textbf{A:} The generative heavy-tailed PLDA model can be used to generate synthetic data, with known properties. We found this invaluable in experiments to explore various discriminative training criteria.

\section{Acknowledgements}
This work was started at the Johns Hopkins University HLTCOE SCALE 2017 Workshop. The authors thank the workshop organizers for inviting us to attend and (in the case of Niko Br\"ummer) for generous travel funding. Themos Stafylakis has been funded by the European Commission program Horizon 2020, under grant agreement no.706668 (Talking Heads). The work was also supported by Czech Ministry of Interior project No. VI20152020025 ``DRAPAK'' Google Faculty Research Award program,  Technology Agency of the Czech Republic project No. TJ01000208 ``NOSICI'', and by Czech Ministry of Education, Youth and Sports from the National Programme of Sustainability (NPU II) project ``IT4Innovations excellence in science - LQ1602''.

\bibliographystyle{IEEEtran}
\bibliography{Anya_Odyssey2018}

\end{document}